\documentclass{article} 
\usepackage{iclr2024_conference,times}

\usepackage{amsmath,amsfonts,bm}

\def\eqref#1{equation~\ref{#1}}

\def\1{\bm{1}}

\def\vk{{\bm{k}}}

\def\vq{{\bm{q}}}

\def\vv{{\bm{v}}}

\DeclareMathAlphabet{\mathsfit}{\encodingdefault}{\sfdefault}{m}{sl}
\SetMathAlphabet{\mathsfit}{bold}{\encodingdefault}{\sfdefault}{bx}{n}

\usepackage{hyperref}
\usepackage{url}
\newcommand{\ours}{S2-Attention\xspace}

\usepackage{latexsym}
\usepackage{algorithm}
\usepackage{algpseudocode}
\usepackage{subcaption}
\usepackage{graphicx}
\usepackage{subcaption}
\usepackage{amsmath}

\usepackage{natbib}
\usepackage{array}
\usepackage{tabularx}
\usepackage{booktabs}
\usepackage{wrapfig}

\usepackage[utf8]{inputenc} 
\usepackage[T1]{fontenc}    
\usepackage{hyperref}       
\usepackage{url}            
\usepackage{booktabs}       
\usepackage{amsfonts}       
\usepackage{nicefrac}       
\usepackage{microtype}      
\usepackage{xcolor}         
\usepackage{standalone}
\usepackage{latexsym}
\usepackage{amsmath}
\usepackage{amssymb}
\usepackage{amsthm}
\usepackage{natbib}   
\usepackage{graphicx}
\usepackage{subcaption}
\usepackage{array}
\usepackage{tabu}
\usepackage{makecell}
\usepackage{paralist}
\usepackage{cases}
\usepackage{diagbox}
\usepackage{enumitem}
\usepackage{soul}
\usepackage{multirow}
\usepackage{verbatim}
\usepackage{tabulary}
\usepackage{booktabs}
\usepackage{tabularx}
\usepackage[mathscr]{euscript}
\usepackage{mathtools}
\usepackage{algorithm}
\usepackage{algpseudocode}
\usepackage{stmaryrd}
\usepackage{tikz-dependency}
\usetikzlibrary{automata,decorations.markings,arrows,positioning,matrix,calc,patterns,angles,quotes,calc}
\usepackage{adjustbox}
\usepackage{tabularx}
\usepackage{xspace}
\usepackage{tabulary}
\usepackage{afterpage}
\usepackage{bm}
\usepackage{color}
\usepackage{graphicx}
\usepackage{slashbox}
\usepackage[toc,page]{appendix}
\usepackage{makecell}
\usepackage{boldline}
\usepackage[shortcuts]{extdash}  

\usepackage{blindtext}
\usepackage{graphicx}
\usepackage{capt-of}
\usepackage{booktabs}
\usepackage{varwidth}
\usepackage{pifont}
\usepackage{wrapfig}

\usepackage{listings}

\usepackage{pythonhighlight}

\definecolor{orange}{rgb}{1,0.5,0}
\definecolor{mdgreen}{rgb}{0.05,0.6,0.05}
\definecolor{mdblue}{rgb}{0,0,0.7}
\definecolor{dkblue}{rgb}{0,0,0.5}
\definecolor{dkgray}{rgb}{0.3,0.3,0.3}
\definecolor{slate}{rgb}{0.25,0.25,0.4}
\definecolor{gray}{rgb}{0.5,0.5,0.5}
\definecolor{ltgray}{rgb}{0.7,0.7,0.7}
\definecolor{purple}{rgb}{0.7,0,1.0}
\definecolor{lavender}{rgb}{0.65,0.55,1.0}

\definecolor{mypurple}{RGB}{111,61,121}
\definecolor{myblue}{RGB}{46,88,180}
\definecolor{myred}{RGB}{181,68,106}
\definecolor{myyellow}{RGB}{204,143,55}

\newcommand{\term}[1]{\textbf{#1}} 

\DeclareSymbolFont{extraup}{U}{zavm}{m}{n}
\DeclareMathSymbol{\vardiamond}{\mathalpha}{extraup}{87}

\newcolumntype{L}[1]{>{\raggedright\let\newline\\\arraybackslash\hspace{0pt}}m{#1}}
\newcolumntype{C}[1]{>{\centering\let\newline\\\arraybackslash\hspace{0pt}}m{#1}}
\newcolumntype{R}[1]{>{\raggedleft\let\newline\\\arraybackslash\hspace{0pt}}m{#1}}

\theoremstyle{plain}

\theoremstyle{definition}

\theoremstyle{remark}

\usepackage[textsize=tiny]{todonotes}

\setul{1pt}{.4pt}

\newcommand*{\modelname}{\textsc{HHST}\xspace}
\newcommand*{\kernelname}{\textsc{S2-Attention}\xspace}

\DeclareFixedFont{\ttb}{T1}{txtt}{bx}{n}{12} 
\DeclareFixedFont{\ttm}{T1}{txtt}{m}{n}{12}  

\title{S2-Attention: Hardware-Aware Context \\ Sharding Among Attention Heads}

\author{Xihui Lin$^{1}$\thanks{Leading Authors. Xihui Lin, Yunan Zhang, and Suyu Ge contribute equally. Code is available at https://github.com/linxihui/dkernel}, Yunan Zhang$^{1*}$, Suyu Ge$^{2*}$\AND Liliang Ren$^{1}$, Barun Patra$^{1}$,  Vishrav Chaudhary$^{1}$, Hao Peng$^{2}$, Xia Song$^{1}$\\
$^{1}$Microsoft, $^{2}$UIUC\\
\texttt{\{xihlin,yunanzhang\}@microsoft.com}\\
}

\iclrfinalcopy 
\begin{document}

\maketitle
\begin{abstract}

Sparse attention, which selectively attends to a subset of tokens in the context, has been an established approach to enhance the efficiency of Transformers. 
However, its theoretical reduction in FLOPs has rarely translated into wall-clock speed-up over its dense attention counterparts, mainly due to the lack of hardware-level optimizations like FlashAttention \citep{flash2}.
Meanwhile, it remains unclear whether sparse attention can maintain the model's quality at the scale of today's large language models (LLMs), and how this can be achieved.
This paper presents Sparsely-Sharded Attention (\kernelname), an optimized Triton kernel library providing a variety of customizable sparse attention implementations for both training and inference.
\kernelname allows customizing the attention patterns at per head per context range level.
The fresh insights from \kernelname inspire a novel sparse attention architecture that meets several desiderata that we find crucial for achieving both practical efficiency gains and strong accuracy on downstream tasks, called as Head-Heterogenous Strided Transformer (\modelname).
For higher sparsity, \modelname shards the context heterogeneously across attention heads, where each head attends to a different subset of tokens while collectively covering the whole. 
We evaluate \modelname by pretraining 1.3B and 7B sized models.
For attention computation, \modelname with \kernelname achieves 8.8$\times$ and 15.9$\times$ wall-clock attention speedup, as well as 2.8$\times$ and 2.5$\times$ training time reduction compared to a dense attention baseline implemented with FlashAttention-2.
Moreover, \modelname's downstream task performance is on-par with dense attention, and achieves a perfect retrieval accuracy at a 128K context length at 7B scale.
At inference, our 7B \modelname, achieves a 4.5$\times$ speed-up compared to the dense counterparts in vLLM. 
\kernelname is released with easy-to-customize APIs for direct usage in Megatron and vLLM. 

\end{abstract}

\section{Introduction}
Transformer-based LLMs have opened up fresh opportunities to both research and applications~\citep{openai2023gpt4,touvron2023llama}. 
Their quadratic complexity imposes prohibitive cost in the training and serving these models.
For example, training Llama 2~\citep{touvron2023llama} 70B with a 4K context length on 2T tokens takes 23 days on 2048 A100 GPUs \cite{cost}.
When serving, the model's KV cache consumes 343GB GPU memory with a 32 batch size and 4K context length.
There is an urgent demand for training LLMs efficiently and serving them cost-effectively.

\begin{figure*}[h!]
\includegraphics[width=\textwidth]{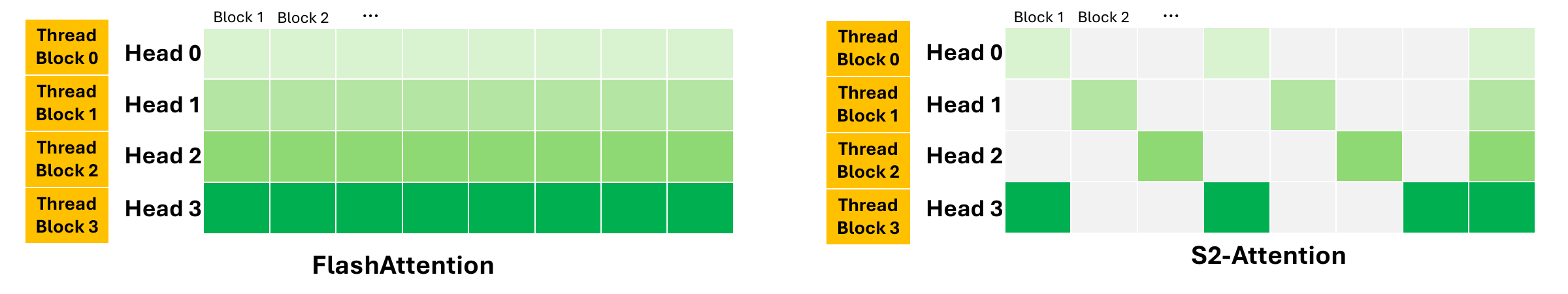}
\caption{Illustration of \ours with four attention heads on a hypothetical GPU with 4 thread blocks. Each attention head is allocated with a shard of the context.}
\label{fig:s2visual}
\end{figure*}
\begin{figure*}[h]
\begin{subfigure}[]{0.6\textwidth}
     \centering
     \includegraphics[width=\textwidth]{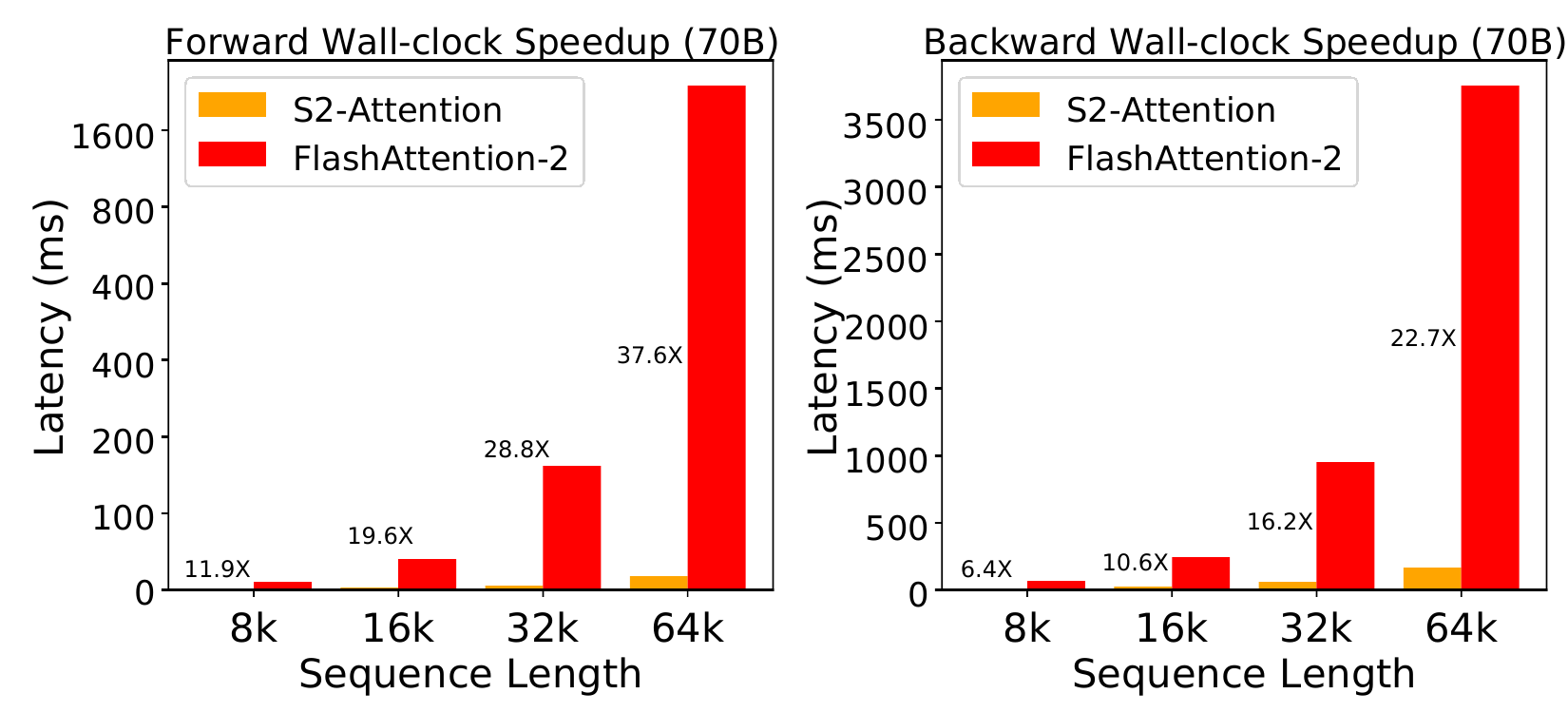}
    \caption{Attention Latency Benchmark over FlashAttention-2.}
\label{fig:attn_bench}
\end{subfigure}
\begin{subfigure}[]{0.35\textwidth}
     \centering
     \includegraphics[width=\textwidth]{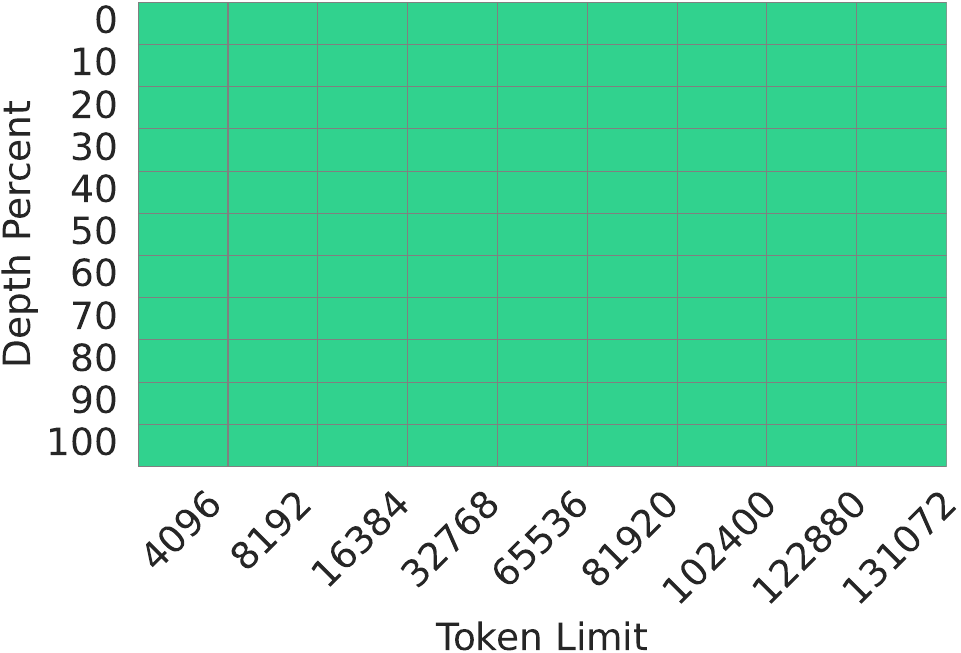}
     \caption{Perfect 128k Needle in a haystack.}
     \label{subfig:helpeval}
\end{subfigure}
\caption{Training Efficiency and long-context analysis of S2-Attention.
Our model, implemented with our kernel, achieves substantial reduction in latency compared to FlashAttention-2 (a). It also achieves perfect retrieval performance at a 128K context length (b).
}
\end{figure*}
Many established works have managed to, at least on paper, improve the efficiency of these models through various \term{sparse attention} techniques~\citep{survey,sparsetransformer,longformer,bigbird},
where only a subset of the tokens in the context are attended to.
However, their theoretical FLOP savings compared to full-context \term{dense attention} often fail to deliver real-world efficiency gains.
As pointed out by the seminal work FlashAttention~\citep{flash,flash2},
the major overhead in attention arises not from computation but from GPU memory access,
especially the shared memory access (SRAM).
Dense attention has benefited from CUDA-level implementations specifically optimized for efficient memory IO, a significant advantage that sparse attention methods have yet to receive.
The absence of a flexible, efficient, and easy-to-use library for optimized sparse attention implementations has become a major roadblock,  delaying progress in both research and applications in improving LLMs' training and serving efficiency.

We aim to bridge this gap with Sparsely-Sharded Attention (\kernelname), 
a Triton library that provides kernel optimization for sparse attention. 
It is highly flexible, allowing practitioners to explore various sparse attention strategies and customize different attention patterns across attention heads and context ranges.
Building a general-purpose fused kernel for sparse attention presents substantial challenges.
In sparse attention, part of the context is not attended to.
As a result, tiling the $Q$, $K$, $V$ tensors, a proven technique that divides large tensors into smaller ones for better parallelization and shared memory (SRAM) usage \citep{flash,flash2}, 
can often result in idling threads and inefficient SRAM usage when the tile size is small.
\kernelname addresses this by efficiently tracking KV usage patterns and dynamically merging query blocks with shared KVs into the same tile. 
This ensures the IO efficiency, regardless of the sparsity granularity, significantly improving SRAM utilization and reducing redundant KV loading.

The insights from the development of \kernelname reveals that \emph{not all sparse attention mechanisms are efficient in practice}.
Many existing training-free sparse attention, including KV eviction methods such as LongGen~\citep{fastgen}, H2O~\citep{h2o}, and MInference~\citep{minfer}, are less compatible with foundational serving mechanisms like continuous batching \citep{continuous}, PagedAttention \citep{pagedatt}, 3D parallelism \citep{megatron}.
For example, in PagedAttention \citep{pagedatt}, evicting tokens from KV blocks would only increase internal fragmentation, and bring extra overhead in scheduling, which in turn hurts serving throughput.
Meanwhile, recent studies  show that 
training free sparse attention would hurt model's long context capabilities \citep{duoattn,longgen,han-etal-2024-lm}.
This has become a primary reason  why they have limited adoption in industry serving and opens-source inference systems to date~\citep{pagedatt,sglang}.

These fresh insights lead to Head-Heterogenous Strided Transforme (\modelname), a new sparse attention approach following key design principles (\S\ref{subsec:kvdiscussion}), which we find crucial for achieving efficiency gains in practice while maintaining strong accuracy on downstream tasks:
\begin{itemize}[noitemsep,topsep=0pt,parsep=0pt,partopsep=0pt]
    \item[(1)] \modelname is  designed with hardware and software systems in mind.
    It applies a novel hardware-friendly sharding strategy across attention heads, where each head attends to a distinct set of tokens following a strided pattern, while collectively covering the context in full (Figure~\ref{fig:s2visual}; \S\ref{subsec:shardformulation}).
    \item[(2)] In order to achieve strong performance on challenging long-context tasks, it is crucial to include direct access to all tokens, at least at certain layers. \modelname achieves this with a hybrid architecture that combines sparse and dense attention across layers, and balances efficiency and performance (\S\ref{subsec:hybrid}).
\end{itemize}

\kernelname is applicable in both training and inference, substantially lowering the barrier to exploring novel sparse attention architectures, which we explore in \S\ref{sec:formulation} and \S\ref{sec:exp}.
We pretrain a suite of models at 1.3B, 7B scales with different sparse attention, and compare them to the dense attention baseline.
Our results show that our \modelname-7B matches the performance of dense attention while achieving a 2.5$\times$ training speed-up and 4.5$\times$ inference speed-up.
Moreover, we extend the 1.3B models to a 32K context length, and 7B models to 128K.
We show that our \modelname can achieve perfect Needle in a Haystack retrieval \citep{kamneedle}.
Compared to FlashAttention-2~\citep{flash2}, \modelname can achieve 8.8$\times$ and 15.9$\times$ attention speed-up for 1.3B, 7B scales, and 2.8$\times$, 2.5$\times$ training wall-clock time reduction.

\kernelname is compatible with commonly used LLM frameworks including PyTorch, Megatron, HuggingFace, and vLLM.
With its user-friendly APIs, importing and customizing \kernelname take no more than several lines of code as shown in Appendix B.

\section{Related Works}
We present our analysis and observations on existing sparse attention attempts in both training and inference.
\begin{figure}[h!]
\centering
\includegraphics[width=0.6\textwidth]{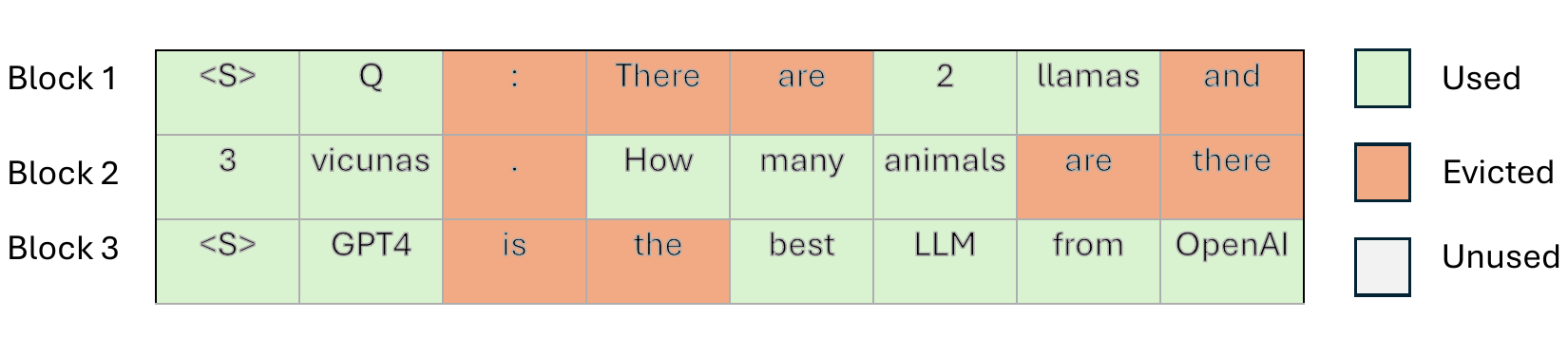}
\caption{Illustration of why KV eviction methods can cause more fragmentation. Here we show 3 pages of KV blocks containing 2 requests. Despite many tokens were evicted, the released slots can hardly be utilized by other requests, leading to higher rate of internal fragmentation.
}
\label{fig:fragmentation}
\vspace{-0.5cm}
\end{figure}
\subsection{Absence of Efficient Sparse Attention Kernel}
\label{subsec:absence}
There have been attempts to reduce the computational complexity of attention by only attending to a subset of tokens~\citep{sparsetransformer,lineartrans,reformer,bigbird,longformer}.
However, these methods can't bring wall-clock speed-up in training due to the negligence of realistic memory access cost~\citep{flash}. 
\cite{flash} breaks down the attention computation into smaller block-wise computation to reduce the IO between SRAM and the high bandwidth memory (HBM). 
The hardware implementation of FlashAttention family \citep{flash,flash2} make them the most widely-adopted attention acceleration framework. 
It remains unclear whether we can implement various sparse self-attention in such hardware-aware way, so that the training speed can be further boosted over FlashAttention.
\subsection{Issues with Plug-in-and-Play KV Eviction Methods}
Recently, plug-in-and-play KV eviction works thrives. 
More specifically, these methods dynamically drop KV vectors at inference to reduce the memory footprint based on certain criteria that designed to preserve model capabilities.

However, we observe such designs are hardly compatible with existing serving systems, which relies on PagedAttention and continuous batching for efficient memory management.
As shown in Figure~\ref{fig:fragmentation}, during KV eviction, the corresponding tokens are release from the physical memory.
However, as the token-wise eviction are not guaranteed to be contiguous, the released memory slots can not be effectively allocated for other requests, known as internal fragmentation.
In this example, the internal fragmentation increases by 37.5\%, which in turn hurts throughput.

Meanwhile, dynamic eviction also introduces overheads in scheduling.
For example, if different heads have a different eviction policy/rate, the faster heads will have to wait for the slower ones, which is a classic load-unbalance scenario.
The issue is more severe when serving larger models, where computations are distributed across devices and nodes with tensor parallel and pipeline parallel.
Such drawbacks further prevent these algorithms from being integrated into real-world serving clusters with hundreds of nodes.

\begin{figure*}[t]
\includegraphics[width=0.9\textwidth]{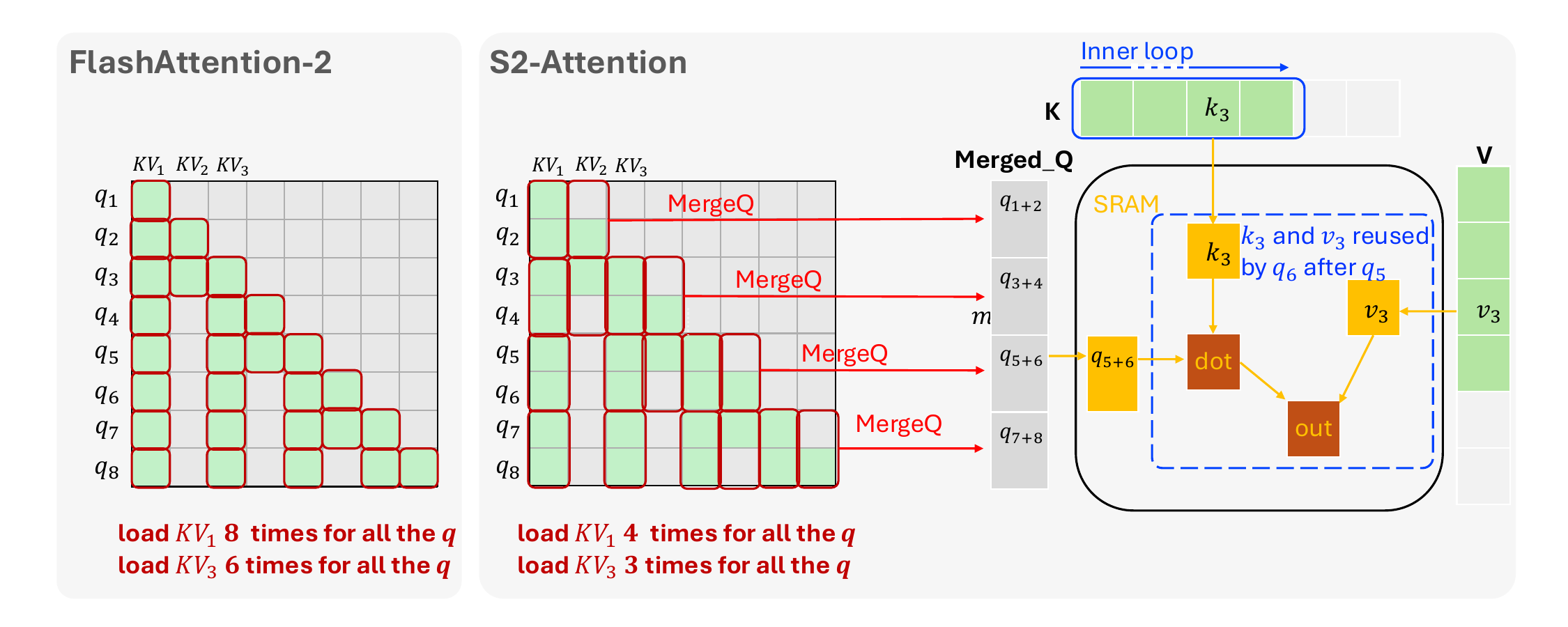}
\caption{Illustration of \ours Implementation. Left: Directly apply FlashAttention-2 tiling to sparse attention. Right: MergeQ, which adaptively merge queries sharing the same KV together when loading into the SRAM, thus reduce redundant KV loading and improve IO efficiency.
}
\label{fig:system}
\end{figure*}

\subsection{Performance Degradation}
Existing studies points out both the training and training-free sparse attention methods have performance degradation compared to their dense counterparts, especially in long-context tasks.
Furthermore, we also observe that some training-free methods\citep{minfer,quest} need benchmark-specific hyper parameters to maintain model quality.
When applied to unseen requests, the same method can display unpredictable behavior.
However, in real-world deployment, user queries often have long tail distribution.
Thus, it's not feasible to pre-determine hyper-parameters for unseen user queries, which makes the deployment of such methods risky.

We discuss handling of these observations in sections below.

\section{S2-Attention: Efficiency and Customization}
This section presents \kernelname.
We first briefly review the basics of GPU memory and execution hierarchy, and then introduce our \textbf{Merge-Q} technique, which significantly improves the kernel's efficiency while allowing more fine-grained customization of the sparse attention.

\subsection{Preliminaries}\label{subsec:preliminaries}
GPU threads have access to a hierarchy of different types of memory.
Global high-bandwidth memory (\textbf{HBM}) is the slowest but largest (roughly $>100\times$ in latency and $\sim6K\times$ in size).
Shared memory (\textbf{SRAM}) is physically on chip, thus has larger bandwidth and lower latency compared to HBM.
Optimizing the computation of the SRAM and minimizing the IO between HBM and SRAM are crucial for improving the efficiency of attention~\citep{flash}. 

Poorly-optimized implementations of attention can result in frequent I/O to HBM and significantly hurt the efficiency.
CUDA organizes threads into \textbf{thread blocks}, which are further divided into warps, groups of 32 threads.
Threads within a block share the data through SRAM.
It is desirable that different threads in the same warp take the same execution path since otherwise efficiency will be hurt due to warp divergence.
Besides, thread block size should be sufficiently large to achieve good utilization and load balancing.
A \textbf{tile} is a portion of the $Q$, $K$, $V$ tensors assigned to a thread block to be processed.
For clarity, we take tile size as block size.
FlashAttention improves efficiency by minimizing HBM I/O, tiling the $Q$, $K$, $V$ tensors into chunks that fit into SRAM for efficient computation \citep{flash}, a principle that this work follows.

\subsection{\kernelname}\label{subsec:s2}
\begin{figure*}[t]
\centering
\includegraphics[width=0.9\textwidth]{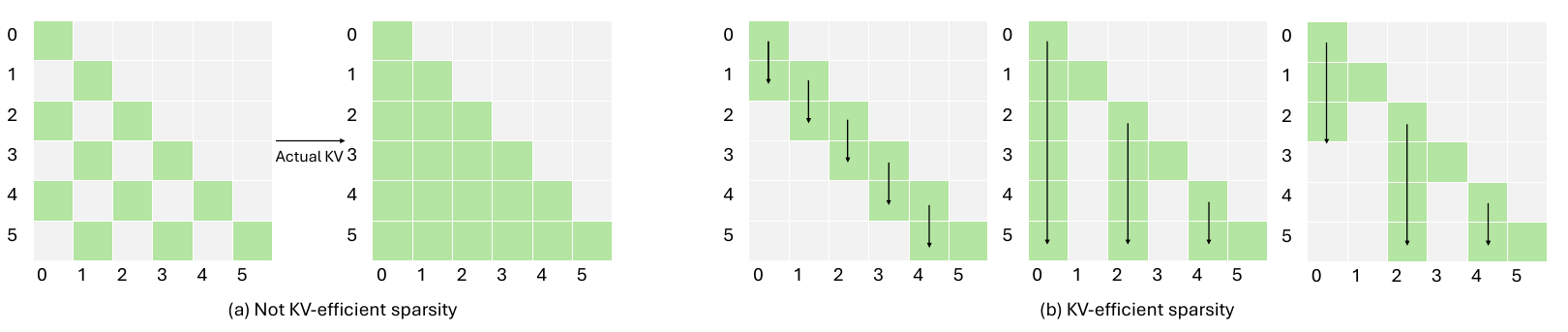}
\caption{
\textbf{(a):} The dilated attention based on relative position as an example of sparse attention that is not KV-efficient. E.g., step 5 attends to KV at positions 1, 3, 5, while step 4 attends to 0, 2, 4.
This results in requiring full KV cache.
Although it suggests nearly 50\% memory savings on paper, it actually requires storing the full KV cache in practice.
\textbf{(b)} All these attention patterns are KV-efficient, as they get pushed to KV-cache when first encountered at decoding, then continuously being attended for several steps before it finally gets evicted (e.g., all tokens in left figure, and token 0 in right figure) and never gets attended again, or remained attended for all future tokens (e.g., tokens 0, 2, 4 in middle figure and tokens 2, 4 in right figure). The arrows show that they all share a "vertical line" pattern.
}
\label{fig:kvefficient}
\end{figure*}

\paragraph{Warmup  (Figure\ref{fig:system} left)}
We first review a simple blocksparse implementation using the FlashAttention algorithm.
A sequence of $N$ tokens is segmented into $B=\lceil N/S\rceil$ shards, each containing $S$ consecutive tokens.
We use $Q_{[i]}$ to denote the query vectors for the $i$th query shard,
and similarly $K_{[i]}$ the key vectors for $i$th key shard.
Following \citet{flash},
for each query vector $q$, we iterate through the $K$ tiles in SRAM to compute $\operatorname{softmax}(qK^\top)$.
Unlike dense attention that uses the entire $K$ tensor, we only consider a subset of keys specified by a sparse attention mask $M$,
which can be stored in a Compressed Sparse Row (CSR) format for memory efficiency.\footnote{\url{https://docs.nvidia.com/nvpl/_static/sparse/storage_format/sparse_matrix.html}}

To better understand the efficiency of such implementation, we can calculate the number of loading needed for each key/value shards.
As shown in Figure \ref{fig:system} (left), the first key/value shards, $KV_1$, is attened by all the query shards, $q_1-q_8$.
Thus, $KV_1$ is loaded 8 times from HBM to SRAM.
If we double the shard size, the number of query shards attending $KV_1$ will be halved to 4.
In this case, $KV_1$ only needs 4 loading which is more efficient.
However, the IO efficiency comes at the cost of granularity of our sparse mask, as we now have to mask-or-keep $2S$ tokens instead of $S$. 
We then discuss how to achieve both IO efficiency and small mask granularity with Merge-Q.

\paragraph{Merge-Q}
\label{subsec:mergeq}
At a high level, the core idea is to merge the query shards attending the same $KV$ blocks  into a single tile so that we don't need separately load the same KV blocks.
In this way, even if the mask granularity becomes smaller, we can still maintain IO efficiency.
Figure \ref{fig:system}: right display a simpler case, where we merge the neighboring two query shards.
Compared to the FlashAttention-2 baseline, this implementation only needs to load $KV_1$ 4 times instead of 8 times with the same mask granularity. 
Merge-Q helps \kernelname support shard sizes as small as 16, enabling a broader range of sparse attention patterns.
Similar ideas can also be applied to merge KV blocks to further boost efficiency.  
We leave more detailed implementation discussion in the released code and Appendix D. 

With \kernelname, the community can customize fine-grained sparse attention patterns with wall-clock speed-up.
However, it remains unclear what types of sparse attention can achieve speed-up without hurting the quality.
We aim to answer this question in the following section.

\section{S2-Attention: Insights, Formulation, and Sparsity Cookbook}
\label{sec:formulation}
In this section, we first discuss which kind of sparse attention patterns allow efficient kernel implementations in practice~(\S\ref{subsec:kvdiscussion}).
Building on these insights, we introduce Head-Heterogenous Strided Transformer (\modelname),
a novel sparse attention architecture (\S\ref{subsec:shardformulation}).

\subsection{KV-Efficient Sparsity}
\label{subsec:kvdiscussion}

KV cache is a primary memory bottlenecks for decoder-only LMs at inference time. 
Many existing sparse attentions determine which tokens to attend to based on relative distances~\citep{sparsetransformer,bigbird,longformer}.
However, these approaches are not GPU memory-efficient during decoding, making it difficult to translate their FLOP savings into real-world efficiency gains.
Figure \ref{fig:kvefficient}(a) provides an illustrative example.
The main issue is that, for such sparse attention, KV not used in earlier decoding steps might be required in later ones, making memory management more challenging.
Despite the nearly 50\% memory saving on paper, it actually requires storing the full KV cache in practice, resulting in zero memory savings.



In contrast, Figure \ref{fig:kvefficient}(b) illustrates a sparse attention that can achieve memory saving in practice.
The key is that the stored KV cache is reused across several decoding steps but is no longer needed in future steps, and thus can be evicted, freeing up the GPU memory. 


The comparison between these two approaches leads to the following 
rule of thumb
of designing KV-efficient sparse attention. For $\forall j \geq i,\, l \geq 1$,
\begin{equation} \label{def:kvefficient}
\begin{aligned}
    &(\vk_i, \vv_i) \mbox{ is attended by } \vq_{j+l} \\
    & \implies (\vk_i, \vv_i) \mbox{ must also be attended by } \vq_j.
\end{aligned}
\end{equation} 
Otherwise,  $\vk_i$ and $\vv_i$ need to be stored at step $j$ for future generations, even it is not used at step $j$. 
Intuitively, in the attention pattern matrix, we shall see continuous "vertical lines" as shown in Figure \ref{fig:kvefficient}(b). This means the sparse patterns should be based on absolute positions rather than relative ones, except for consecutive local context (e.g., left figure in Figure \ref{fig:kvefficient}(b)). 

\subsection{Head-Heterogenous Strided Transformer}
\label{subsec:shardformulation}
This section introduces Head-Heterogenous Strided Transformer (\modelname), a novel efficient sparse attention inspired by the insights we learned above.
Core to its design are two design choices introduced below.

\paragraph{Heterogeneous Context Sharding across Attention Heads}
To achieve balanced load across attention heads and enhance parallelization, each head should attend to an equal number of tokens.
Additionally, \modelname ensures that different heads attend to different shards of the context while collectively covering the entire context. 
This design makes sure that \modelname always has direct access to the full context at each layer, without compromising parallelization.
Figure~\ref{fig:s2visual} provides an illustrative diagram. 

More formally, for context with $B$ shards, 
we take the most recent $B_{l}$ blocks as local blocks and set the rest as remote blocks.
For attention head with index $h$, its $B \times B$ block attention mask $M^h$ is:
\begin{equation*} \label{local-stride-hetero}
M^h_{i,j} = 
\left\{
\begin{array}{lll}
    1, & i-j < B_l, & \text{Local} \\
    1, &  j - o_h \in s \mathbb{Z}_{\geq 0}\,\wedge\, i - j \in [B_l, B) &\text{Stride} \\
    0 & \text{otherwise}
\end{array}
\right. 
\end{equation*}
 $s$ is the stride size, and $x \in m\mathbb{Z}_{\geq 0}$ mean $x$  is 0 or a positive multiple of $m$.
 Similarly to a sliding window, tokens beyond the $B$ shards are not attended to.



The flexibility of our \kernelname kernel enables an efficient implementation of this strategy. As shown in our experiments, this design allows the model to achieve strong long-context performance while maximizing efficiency gains.

\paragraph{Hybrid Architecture}\label{subsec:hybrid}
As previous studies show~\citep{pyramid,lieber2403jamba}, some attention layers are significantly denser compared to the others, with attention weights distributed near uniformly across all positions. 
Therefore, it is particularly beneficial to retain dense attention in these layers.
This motivates us to explore a hybrid architecture that combines our efficient sparse attention in most layers with dense attention in others.
We empirically find that our sparse attention strategy is highly effective, requiring only 1/6 of the attention layers to be dense to achieve strong retrieval performance with 128K-long contexts.
More exploration is presented in our experiments.

\paragraph{Discussion}
It is important to point out that all eviction strategies targeting inference \citep{h2o,scissorhands,fastgen} are  KV-cache efficient, since evicted KV will never be used by future queries. 
However, these strategies introduce sample-dependent sparsity patterns, making it computationally expensive to determine eviction timing during decoding.
In contrast, our approach uses a fixed sparsity pattern across all samples, eliminating the overhead of deciding which tokens to evict.
Besides, KV eviction approaches are post-hoc and often perform much poorly as compared to the original dense counterpart~\citep{longgen}.
Our \modelname, as we will soon see in the experiments, adapts to the sparse attention during training (pre-training or post-training) 
performs comparably to dense baselines while reducing the training overhead.


\section{Experiment}
\label{sec:exp}
To evaluate \modelname,
we first study the pre-training quality in \S\ref{subsec:downstream} and \S\ref{subsec:needle}.
We then benchmark the kernel efficiency and end-to-end serving latency in \S\ref{subsec:attnbench} and \S\ref{subsec:inference}.
Lastly, we conduct an ablation study of the design choices.

\subsection{Benchmarking Model Training Quality}
\label{subsec:downstream}
\paragraph{Settings} We first train a range of 1.3B model with the Llama 2 architecture, with 24 layers, 2048 hidden size with 16 heads, with max sequence length as 8192.
We use the open-source FineWeb-Edu-350B \cite{fineweb} as the pre-training corpus. 
An OpenAI Tiktoken tokenizer with 100K vocabulary size is used to process the raw text. 
All model variations use batch size of 4M tokens for all sequence lengths and train for a total of 300 billion tokens.  
For hyperparameters, we use $\mu$P \cite{mup} with a base shape of 256.
A $\mu$P learning rate of 0.02 is used with linear decay and 0.5\% of total training tokens for warmup.
All models are evaluated after training on the total 300B tokens for one epoch.



\paragraph{Downstream Tasks} We use a model with dense attention as our baseline, denoted as ``Dense''.
To study our hybrid structure with heterogeneous sharding and union completeness, we control the FLOPs to be approximately equivalent.
The total attended tokens is around 576 tokens, or 9 shards of 64 tokens.
We use this to configure the sliding window attention (SWA), as the control set.
We add different changes to SWA to see how they affect the training quality.
The treatment sets are grouped into 1) \textbf{Homogeneous} (Different heads attend to the same shards);
2) \textbf{Heterogeneous \& Incomplete} (Different heads attend to different shards but not covering the entire context), 
and 3) \textbf{Heterogeneous \& Complete} (Different heads attend to the same shards and collectively cover the entire context).

\begin{table*}[!ht]
\centering
\label{tab:downstream}
\caption{Pre-Training quality evaluation. ``SWA'' refers to sliding window attention. ``L'' refers to number of local blocks. ``V'' refers to the vertical stride size. ``+ Sink'' refers attending to attention sink. ``+ Dense'' refers to making the first two attention layers dense.}
\label{tab:downstream}
\begin{adjustbox}{max width=0.8\textwidth}
\begin{tabular}{lrrrrr}\toprule
\textbf{Model}  & \textbf{Passkey} & \textbf{WinoGrande} & \textbf{PIQA} & \textbf{RACE} & \textbf{Wikitext103(ppl)} \\\midrule
\textbf{Dense (Upper Bound)}  & \textbf{0.865} & \textbf{0.592} & \textbf{0.733} & \textbf{0.403} & \textbf{15.884} \\
\midrule
\multicolumn{6}{c}{\textbf{Homogeneous} (18\% FLOPs of Dense)} \\
\midrule
\modelname-L9 (SWA) & 0.334 & 0.547 & 0.705 & 0.363 & 21.997 \\
\modelname-L9 + Dense  & 0.620 & 0.575 & 0.714 & 0.373 & 20.450 \\
\modelname-L9 + Sink & 0.560 & 0.566 & 0.721 & 0.380 & 21.037 \\
\modelname-L9 + Sink + Dense & 0.771 & 0.577 & 0.728 & 0.388 & 18.503 \\
\modelname-L1V15  & 0.542 & 0.541 & 0.716 & 0.352 & 21.035 \\
\modelname-L1V15 + Dense  & 0.741 & 0.568 & 0.713 & 0.349 & 20.579 \\
\midrule
\multicolumn{6}{c}{\textbf{Heterogeneous \& Incomplete} (18\% FLOPs of Dense)} \\
\midrule
\modelname-L2V18 & 0.630 & 0.565 & 0.728 & 0.357 & 20.502 \\
\modelname-L2V18 + Dense  & 0.823 & 0.587 & \textbf{0.732} & 0.379 & 18.726 \\
\modelname-L4V25 & 0.612 & 0.542 & 0.720 & 0.352 & 20.875\\
\modelname-L4V25 + Dense & 0.795 & 0.569 & 0.724 & 0.386 & 19.285 \\
\midrule
\multicolumn{6}{c}{\textbf{Heterogeneous \& Complete} (18\% FLOPs of Dense)} \\
\midrule
\modelname-L1V15 & 0.782 & 0.571 & 0.724 & 0.361 & 19.551 \\
\textbf{\modelname-L1V15 + Dense (\modelname)} &\textbf{ 0.941} & \textbf{0.587} & 0.725 & \textbf{0.397} & \textbf{17.183} \\
\bottomrule
\end{tabular}
\end{adjustbox}
\end{table*}

From Table \ref{tab:downstream}, we can observe the hybrid architectures shows promising results. 
As we can see from \textbf{S2-L1V15 + Dense (\modelname)} in the last row, heterogeneous sharding with complete context and two dense layers give consistently best results across tasks, with minor  gap from the default attention baseline while using only \textbf{18\%} FLOPs.
Notably, in the Passkey Retrieval task, \modelname can achieve much better performance compared to the dense model.
This observation works as an initial validation of the context understanding ability of the \modelname design.
We'll further validate it in the long context continual training section.

We also found adding two dense layers generally leads to a significantly higher performance.
Within the \textbf{Homogeneous} group, we can observe adding attention sink can significantly boost training quality, compared to only using the sliding window (SWA).
In the \textbf{Heterogeneous \& Incomplete} group, the vertical stride size is bigger than the number of attention heads, making the context incomplete after the union.
For the \textbf{Heterogeneous \& Complete} group, we tune the stride size and local window so that it just covers the full context while having the same FLOPs as others.
When comparing the \textbf{Incomplete} group to the \textbf{Complete} group, we can see the benefits of making the union of context complete by limiting vertical stride size.

\subsection{Long Context Continual Training}
\label{subsec:needle}
\begin{figure*}[hbt!]
  \centering
  \begin{subfigure}[b]{0.32\textwidth}
      \centering
      \includegraphics[width=\textwidth]{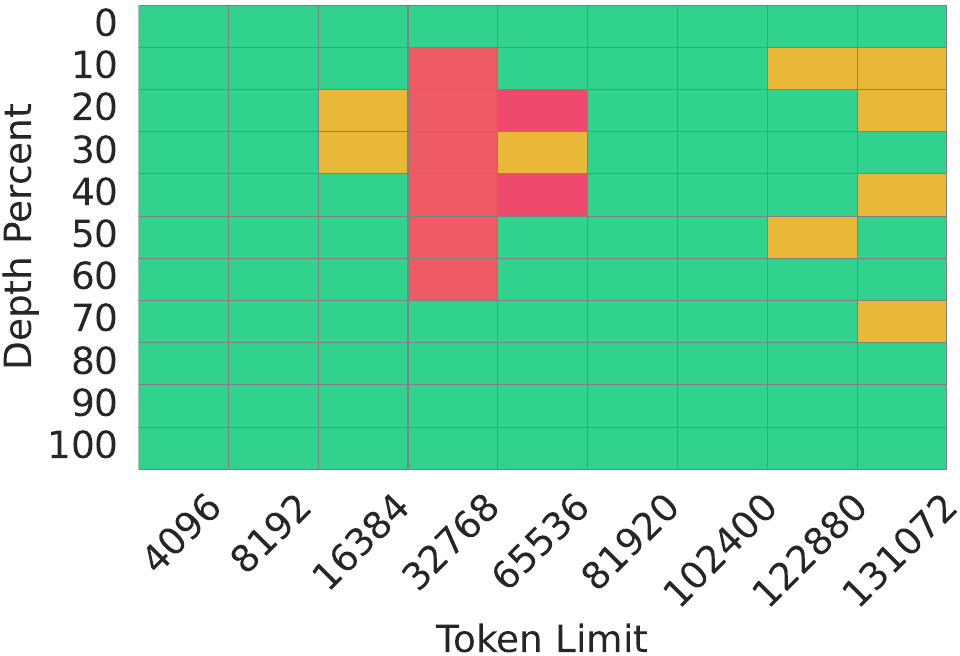}
      \caption{2 dense layers (11\% FLOPs)}
      \label{fig:mergeqfwd}
  \end{subfigure}
  \hfill
  \begin{subfigure}[b]{0.32\textwidth}
      \centering
      \includegraphics[width=\textwidth]{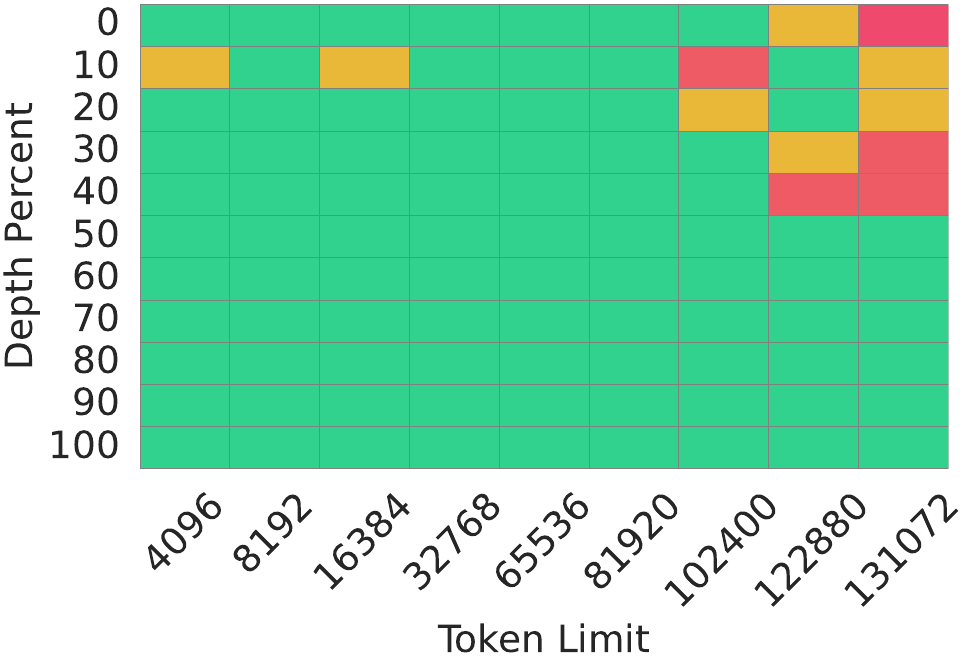}
      \caption{4 dense layers (17\% FLOPs)}
      \label{fig:mergeqbwd}
  \end{subfigure}
  \hfill
  \begin{subfigure}[b]{0.32\textwidth}
      \centering
      \includegraphics[width=\textwidth]{Figures/l31v32dense12-20.pdf}
      \caption{8 dense layers (28\% FLOPs)}
      \label{fig:dsplitfwd}
  \end{subfigure}
  \hfill
  \caption{128K Needle In A Haystack Evaluation. We modify the number of dense layers and demonstrate FLOPs saving over Dense (all layers are dense).}
  \label{fig:128kneedle}
\end{figure*}

We further examine how to adapt sparse attention to longer contexts.
We start from an existing densely pre-trained model and extend its context length by continually training it on a longer context length with \modelname sparse architecture.
Specifically, we choose Llama-2-7B and continually train it on 128K context length.
We change the RoPE base to 5M.
Both models are continually trained with 10B tokens following the recipe in \citet{dataengineer}.
We evaluate the models on the Needle In A Haystack task~\cite{kamneedle}.

To investigate how to achieve strong long-context performance, we modify the number of dense layers in \modelname.
We set the number of dense layers as 2, 4 and 8, respectively.
We fix the number of local blocks to be 31 and vertical stride size to be 32.
As shown in Figure~\ref{fig:128kneedle}, for 128K context, the model can retrieve the full context with 8 dense layers but fails to do so with only 2 and 4 dense layers.
The results validate the long context capability of \modelname design.

\subsection{Training Speed-up}
\label{subsec:attnbench}
\subsubsection{Attention Operation Benchmark}
\textbf{Benchmark Settings} We measure the attention runtime of \modelname with our \kernelname kernel, and FlashAttention-2 on an A100 80GB GPU for different context length, number of head, and head dimension settings.
\begin{figure}[h!]
  \centering
  \begin{subfigure}[b]{0.49\textwidth}
      \centering
      \includegraphics[width=\textwidth]{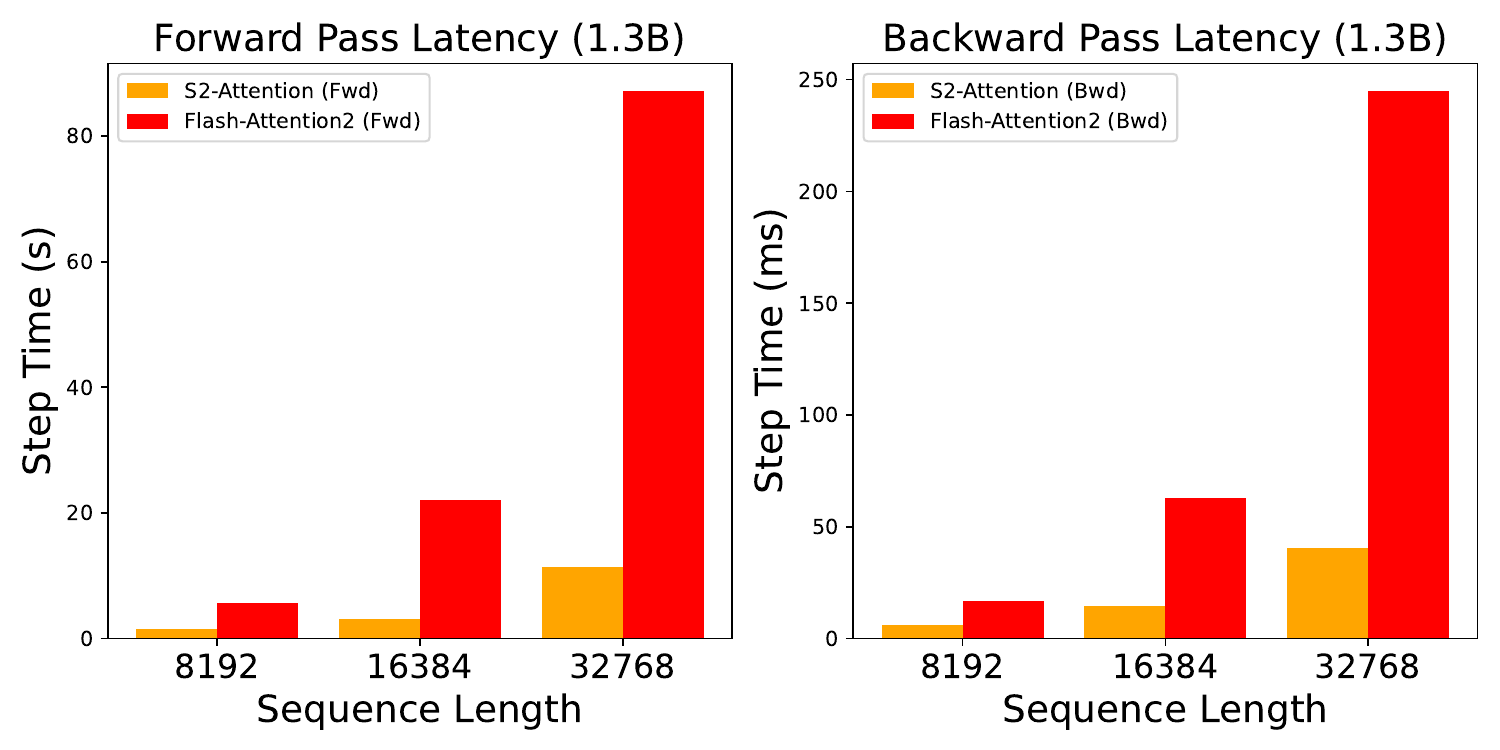}
      \caption{1.3B Attention Speed-up.}
      \label{fig:attn1b}
  \end{subfigure}
  \hfill
  \begin{subfigure}[b]{0.49\textwidth}
      \centering
      \includegraphics[width=\textwidth]{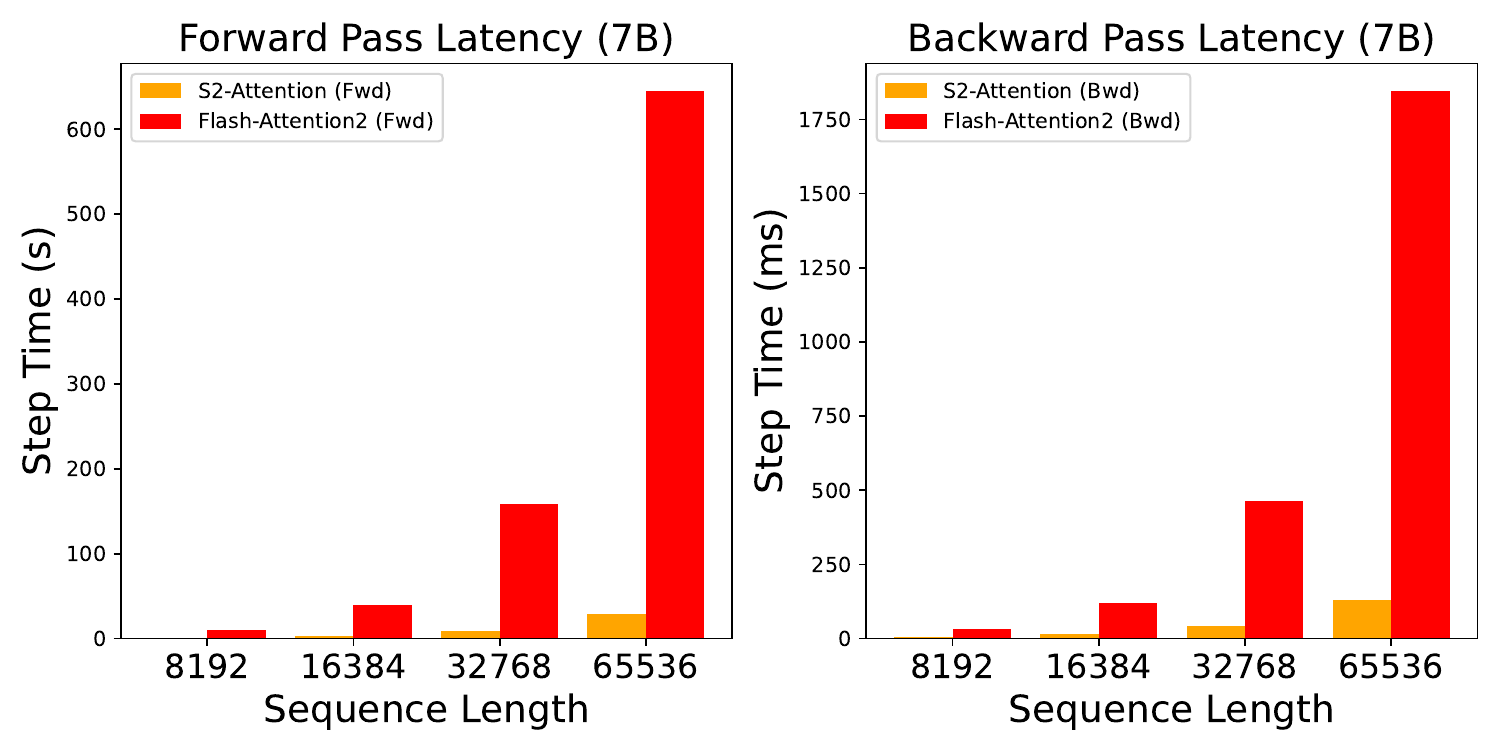}
      \caption{7B Attention Speed-up.}
      \label{fig:attn7b}
  \end{subfigure}
  \caption{Attention Speed-up vs Sequence Length and Model Scale. }
\label{fig:attn_small_bench}
\vspace{-0.5cm}
\end{figure}

In Figure \ref{fig:attn_small_bench} and Figure \ref{fig:attn_bench} We benchmark the speed-up brought by \modelname in 1.3B, 7B, 70B model sizes across different sequence lengths to showcase the scalability of our system.
For all the model sizes, \modelname  can achieve multiple times of speed-up over FlashAttention-2. 
For 70B models with 64 heads, \modelname  can give 25.3$\times$ end-to-end speed-up.
For example, in 1.3B models with a vertical stride of 16, \modelname  can achieve a 8.8$\times$ speed-up. 
As the max sequence length grows longer, the speed-up gradually approximates the theoretical FLOPs reduction benefits.
The overall boost is hedged a bit due to our less optimized backward kernel, which leaves room for further improvement.
\subsection{Training and Inference Speed-up}
\label{subsec:attnbench}
\begin{figure}[h!]
  \centering
  \begin{subfigure}[b]{0.4\textwidth}
      \centering
      \includegraphics[width=\textwidth]{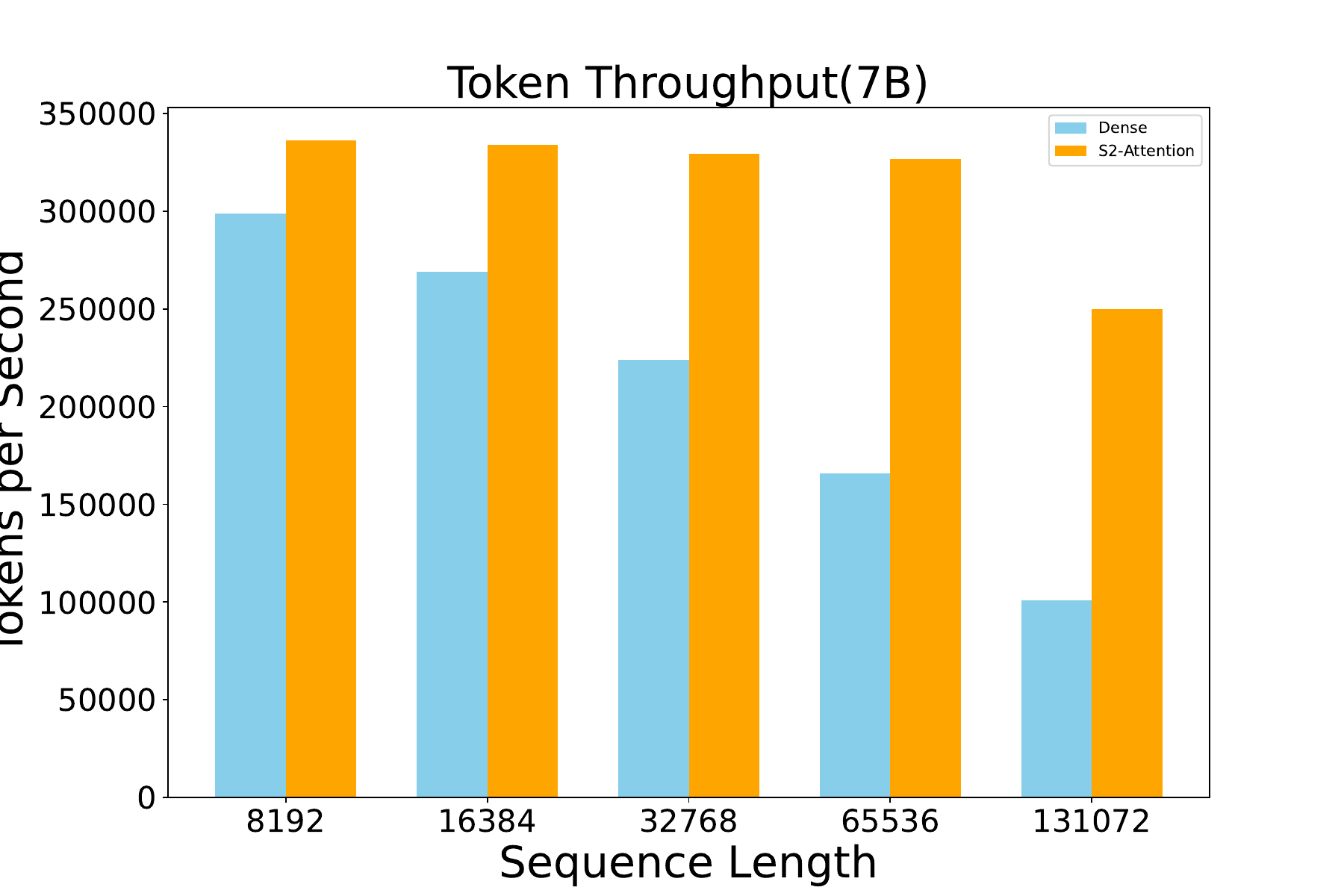}
      \caption{7B end-to-end training speed-up.}
      \label{fig:attn7btrain}
  \end{subfigure}
  \hfill
  \begin{subfigure}[b]{0.5\textwidth}
      \centering
      \includegraphics[width=\textwidth]{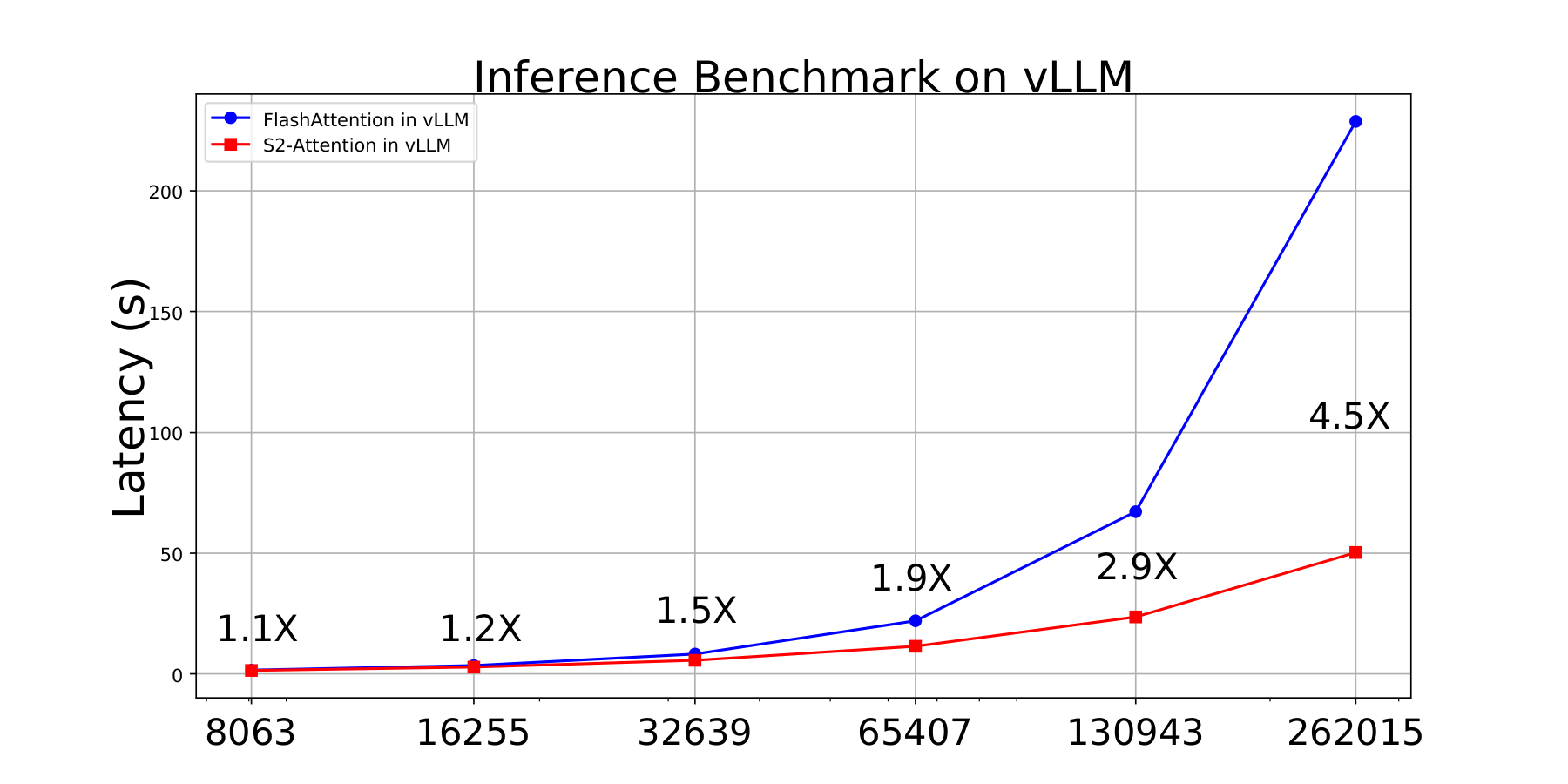}
      \caption{7B end-to-end inference speed-up on vLLM.}
      \label{fig:infer_prefill}
  \end{subfigure}
  
\label{fig:e2e_bench}
\end{figure}
We evaluate the end to end training speed-up of the 1.3B and 7B models by measuring the token throughput of both models. 
All models are trained on 256 A100, with a batch size of 8M tokens and activation checkpointing.
For 1.3B, \modelname  can get 1.2$\times$, 1.8$\times$, 2.3$\times$, and 2.8$\times$ token throughput on 8K to 128K context compared to FlashAttention-2.
For 7B models, \modelname  can get 1.1$\times$, 1.2$\times$, 1.5$\times$, and 2.5$\times$ token throughput improvement.
In order to demonstrate the inference efficiency improvements of \modelname, we measure the end-to-end latency over different context length settings.
To make comparison realistic, our experiments are done on vLLM~\citep{pagedatt}.
We choose the FlashAttention-2 backend in vLLM as baseline for fair comparison, as the inference kernel of \kernelname is also based on vLLM.
Both methods are deployed on a single node with 8 A100 80GPU, with tensor parallel size equals 4.
We set output length as 128 and vary input length between 16K to 256K. 
As shown in Figure \ref{fig:infer_prefill}, \modelname can achieves 1.1$\times$, 1.2$\times$, 2.9$\times$, 4.5$\times$ speed-up on 8K, 16K, 128K, 256K context.

\section{Conclusion}
We presented \kernelname, an optimized Triton kernel library that provides a variety of customizable sparse attention implementations for both training and inference.
The insights from \kernelname led to several principles about designing 
choices of sparse attention methods to make them efficiency in practice.
They inspired a novel hybrid sparse attention architecture that meets several desiderata that we find crucial for achieving both practical efficiency gains and strong accuracy on downstream tasks, called as Head-Heterogenous Strided Transformer (\modelname).
We will open-source our kernel library and make it a plug-in-and-play alternative for  FlashAttention-2 module in popular training frameworks like Megatron and Pytorch.
We also integrated \kernelname into vLLM backend for instant serving. 
Both the training and inference kernels allow users to freely customize their sparsity pattern, facilitating the whole community to study the topic in the future.
\bibliography{custom}

\begin{thebibliography}{29}
\providecommand{\natexlab}[1]{#1}
\providecommand{\url}[1]{\texttt{#1}}
\expandafter\ifx\csname urlstyle\endcsname\relax
  \providecommand{\doi}[1]{doi: #1}\else
  \providecommand{\doi}{doi: \begingroup \urlstyle{rm}\Url}\fi

\bibitem[Beltagy et~al.(2020)Beltagy, Peters, and Cohan]{longformer}
Iz~Beltagy, Matthew~E. Peters, and Arman Cohan.
\newblock Longformer: The long-document transformer.
\newblock \emph{CoRR}, abs/2004.05150, 2020.
\newblock URL \url{https://arxiv.org/abs/2004.05150}.

\bibitem[Child et~al.(2019)Child, Gray, Radford, and Sutskever]{sparsetransformer}
Rewon Child, Scott Gray, Alec Radford, and Ilya Sutskever.
\newblock Generating long sequences with sparse transformers.
\newblock \emph{CoRR}, abs/1904.10509, 2019.
\newblock URL \url{http://arxiv.org/abs/1904.10509}.

\bibitem[Dao(2023)]{flash2}
Tri Dao.
\newblock Flashattention-2: Faster attention with better parallelism and work partitioning.
\newblock \emph{CoRR}, abs/2307.08691, 2023.
\newblock \doi{10.48550/ARXIV.2307.08691}.
\newblock URL \url{https://doi.org/10.48550/arXiv.2307.08691}.

\bibitem[Dao et~al.(2022)Dao, Fu, Ermon, Rudra, and R{\'{e}}]{flash}
Tri Dao, Daniel~Y. Fu, Stefano Ermon, Atri Rudra, and Christopher R{\'{e}}.
\newblock Flashattention: Fast and memory-efficient exact attention with io-awareness.
\newblock In Sanmi Koyejo, S.~Mohamed, A.~Agarwal, Danielle Belgrave, K.~Cho, and A.~Oh (eds.), \emph{Advances in Neural Information Processing Systems 35: Annual Conference on Neural Information Processing Systems 2022, NeurIPS 2022, New Orleans, LA, USA, November 28 - December 9, 2022}, 2022.
\newblock URL \url{http://papers.nips.cc/paper\_files/paper/2022/hash/67d57c32e20fd0a7a302cb81d36e40d5-Abstract-Conference.html}.

\bibitem[Fu et~al.(2024)Fu, Panda, Niu, Yue, Hajishirzi, Kim, and Peng]{dataengineer}
Yao Fu, Rameswar Panda, Xinyao Niu, Xiang Yue, Hannaneh Hajishirzi, Yoon Kim, and Hao Peng.
\newblock Data engineering for scaling language models to 128k context.
\newblock In \emph{Forty-first International Conference on Machine Learning, {ICML} 2024, Vienna, Austria, July 21-27, 2024}. OpenReview.net, 2024.
\newblock URL \url{https://openreview.net/forum?id=TaAqeo7lUh}.

\bibitem[Ge et~al.(2024{\natexlab{a}})Ge, Lin, Zhang, Han, and Peng]{longgen}
Suyu Ge, Xihui Lin, Yunan Zhang, Jiawei Han, and Hao Peng.
\newblock A little goes a long way: Efficient long context training and inference with partial contexts.
\newblock \emph{arXiv preprint arXiv:2410.01485}, 2024{\natexlab{a}}.

\bibitem[Ge et~al.(2024{\natexlab{b}})Ge, Zhang, Liu, Zhang, Han, and Gao]{fastgen}
Suyu Ge, Yunan Zhang, Liyuan Liu, Minjia Zhang, Jiawei Han, and Jianfeng Gao.
\newblock Model tells you what to discard: Adaptive {KV} cache compression for llms.
\newblock In \emph{The Twelfth International Conference on Learning Representations, {ICLR} 2024, Vienna, Austria, May 7-11, 2024}. OpenReview.net, 2024{\natexlab{b}}.
\newblock URL \url{https://openreview.net/pdf?id=88nT0j5jAn}.

\bibitem[Han et~al.(2024)Han, Wang, Peng, Xiong, Chen, Ji, and Wang]{han-etal-2024-lm}
Chi Han, Qifan Wang, Hao Peng, Wenhan Xiong, Yu~Chen, Heng Ji, and Sinong Wang.
\newblock {LM}-infinite: Zero-shot extreme length generalization for large language models.
\newblock In Kevin Duh, Helena Gomez, and Steven Bethard (eds.), \emph{Proceedings of the 2024 Conference of the North American Chapter of the Association for Computational Linguistics: Human Language Technologies (Volume 1: Long Papers)}, pp.\  3991--4008, Mexico City, Mexico, June 2024. Association for Computational Linguistics.
\newblock \doi{10.18653/v1/2024.naacl-long.222}.
\newblock URL \url{https://aclanthology.org/2024.naacl-long.222/}.

\bibitem[Huang et~al.(2022)Huang, Khetan, Bidart, and Karnin]{pyramid}
Xin Huang, Ashish Khetan, Rene Bidart, and Zohar Karnin.
\newblock Pyramid-bert: Reducing complexity via successive core-set based token selection.
\newblock In Smaranda Muresan, Preslav Nakov, and Aline Villavicencio (eds.), \emph{Proceedings of the 60th Annual Meeting of the Association for Computational Linguistics (Volume 1: Long Papers), {ACL} 2022, Dublin, Ireland, May 22-27, 2022}, pp.\  8798--8817. Association for Computational Linguistics, 2022.
\newblock \doi{10.18653/v1/2022.acl-long.602}.
\newblock URL \url{https://doi.org/10.18653/v1/2022.acl-long.602}.

\bibitem[Jiang et~al.(2024)Jiang, Li, Zhang, Wu, Luo, Ahn, Han, Abdi, Li, Lin, Yang, and Qiu]{minfer}
Huiqiang Jiang, Yucheng Li, Chengruidong Zhang, Qianhui Wu, Xufang Luo, Surin Ahn, Zhenhua Han, Amir~H. Abdi, Dongsheng Li, Chin{-}Yew Lin, Yuqing Yang, and Lili Qiu.
\newblock Minference 1.0: Accelerating pre-filling for long-context llms via dynamic sparse attention.
\newblock \emph{CoRR}, abs/2407.02490, 2024.
\newblock \doi{10.48550/ARXIV.2407.02490}.
\newblock URL \url{https://doi.org/10.48550/arXiv.2407.02490}.

\bibitem[Kamradt(2023)]{kamneedle}
Greg Kamradt.
\newblock Needle in a haystack-pressure testing llms.
\newblock \emph{Github Repository}, pp.\ ~28, 2023.

\bibitem[Katharopoulos et~al.(2020)Katharopoulos, Vyas, Pappas, and Fleuret]{lineartrans}
Angelos Katharopoulos, Apoorv Vyas, Nikolaos Pappas, and Fran{\c{c}}ois Fleuret.
\newblock Transformers are rnns: Fast autoregressive transformers with linear attention.
\newblock In \emph{Proceedings of the 37th International Conference on Machine Learning, {ICML} 2020, 13-18 July 2020, Virtual Event}, volume 119 of \emph{Proceedings of Machine Learning Research}, pp.\  5156--5165. {PMLR}, 2020.
\newblock URL \url{http://proceedings.mlr.press/v119/katharopoulos20a.html}.

\bibitem[Kitaev et~al.(2020)Kitaev, Kaiser, and Levskaya]{reformer}
Nikita Kitaev, Lukasz Kaiser, and Anselm Levskaya.
\newblock Reformer: The efficient transformer.
\newblock In \emph{8th International Conference on Learning Representations, {ICLR} 2020, Addis Ababa, Ethiopia, April 26-30, 2020}. OpenReview.net, 2020.
\newblock URL \url{https://openreview.net/forum?id=rkgNKkHtvB}.

\bibitem[Kwon et~al.(2023)Kwon, Li, Zhuang, Sheng, Zheng, Yu, Gonzalez, Zhang, and Stoica]{pagedatt}
Woosuk Kwon, Zhuohan Li, Siyuan Zhuang, Ying Sheng, Lianmin Zheng, Cody~Hao Yu, Joseph~E. Gonzalez, Hao Zhang, and Ion Stoica.
\newblock Efficient memory management for large language model serving with pagedattention.
\newblock \emph{CoRR}, abs/2309.06180, 2023.
\newblock \doi{10.48550/arXiv.2309.06180}.
\newblock URL \url{https://doi.org/10.48550/arXiv.2309.06180}.

\bibitem[Lieber et~al.()Lieber, Lenz, Bata, Cohen, Osin, Dalmedigos, Safahi, Meirom, Belinkov, Shalev-Shwartz, et~al.]{lieber2403jamba}
Opher Lieber, Barak Lenz, Hofit Bata, Gal Cohen, Jhonathan Osin, Itay Dalmedigos, Erez Safahi, Shaked Meirom, Yonatan Belinkov, Shai Shalev-Shwartz, et~al.
\newblock Jamba: A hybrid transformer-mamba language model, 2024.
\newblock \emph{URL https://arxiv. org/abs/2403.19887}.

\bibitem[Liu et~al.(2023)Liu, Desai, Liao, Wang, Xie, Xu, Kyrillidis, and Shrivastava]{scissorhands}
Zichang Liu, Aditya Desai, Fangshuo Liao, Weitao Wang, Victor Xie, Zhaozhuo Xu, Anastasios Kyrillidis, and Anshumali Shrivastava.
\newblock Scissorhands: Exploiting the persistence of importance hypothesis for {LLM} {KV} cache compression at test time.
\newblock \emph{CoRR}, abs/2305.17118, 2023.
\newblock \doi{10.48550/arXiv.2305.17118}.
\newblock URL \url{https://doi.org/10.48550/arXiv.2305.17118}.

\bibitem[OpenAI(2023)]{openai2023gpt4}
OpenAI.
\newblock Gpt-4 technical report, 2023.

\bibitem[Penedo et~al.(2024)Penedo, Kydl{\'{\i}}cek, Allal, Lozhkov, Mitchell, Raffel, von Werra, and Wolf]{fineweb}
Guilherme Penedo, Hynek Kydl{\'{\i}}cek, Loubna~Ben Allal, Anton Lozhkov, Margaret Mitchell, Colin Raffel, Leandro von Werra, and Thomas Wolf.
\newblock The fineweb datasets: Decanting the web for the finest text data at scale.
\newblock \emph{CoRR}, abs/2406.17557, 2024.
\newblock \doi{10.48550/ARXIV.2406.17557}.
\newblock URL \url{https://doi.org/10.48550/arXiv.2406.17557}.

\bibitem[Rucinski(2024)]{cost}
Szymon Rucinski.
\newblock Efficient language adaptive pre-training: Extending state-of-the-art large language models for polish.
\newblock \emph{CoRR}, abs/2402.09759, 2024.
\newblock \doi{10.48550/ARXIV.2402.09759}.
\newblock URL \url{https://doi.org/10.48550/arXiv.2402.09759}.

\bibitem[Shoeybi et~al.(2020)Shoeybi, Patwary, Puri, LeGresley, Casper, and Catanzaro]{megatron}
Mohammad Shoeybi, Mostofa Patwary, Raul Puri, Patrick LeGresley, Jared Casper, and Bryan Catanzaro.
\newblock Megatron-lm: Training multi-billion parameter language models using model parallelism, 2020.

\bibitem[Tang et~al.(2024)Tang, Zhao, Zhu, Xiao, Kasikci, and Han]{quest}
Jiaming Tang, Yilong Zhao, Kan Zhu, Guangxuan Xiao, Baris Kasikci, and Song Han.
\newblock Quest: Query-aware sparsity for efficient long-context llm inference, 2024.
\newblock URL \url{https://arxiv.org/abs/2406.10774}.

\bibitem[Tay et~al.(2023)Tay, Dehghani, Bahri, and Metzler]{survey}
Yi~Tay, Mostafa Dehghani, Dara Bahri, and Donald Metzler.
\newblock Efficient transformers: {A} survey.
\newblock \emph{{ACM} Comput. Surv.}, 55\penalty0 (6):\penalty0 109:1--109:28, 2023.
\newblock \doi{10.1145/3530811}.
\newblock URL \url{https://doi.org/10.1145/3530811}.

\bibitem[Touvron et~al.(2023)Touvron, Martin, Stone, Albert, Almahairi, Babaei, Bashlykov, Batra, Bhargava, Bhosale, Bikel, Blecher, Ferrer, Chen, Cucurull, Esiobu, Fernandes, Fu, Fu, Fuller, Gao, Goswami, Goyal, Hartshorn, Hosseini, Hou, Inan, Kardas, Kerkez, Khabsa, Kloumann, Korenev, Koura, Lachaux, Lavril, Lee, Liskovich, Lu, Mao, Martinet, Mihaylov, Mishra, Molybog, Nie, Poulton, Reizenstein, Rungta, Saladi, Schelten, Silva, Smith, Subramanian, Tan, Tang, Taylor, Williams, Kuan, Xu, Yan, Zarov, Zhang, Fan, Kambadur, Narang, Rodriguez, Stojnic, Edunov, and Scialom]{touvron2023llama}
Hugo Touvron, Louis Martin, Kevin Stone, Peter Albert, Amjad Almahairi, Yasmine Babaei, Nikolay Bashlykov, Soumya Batra, Prajjwal Bhargava, Shruti Bhosale, Dan Bikel, Lukas Blecher, Cristian~Canton Ferrer, Moya Chen, Guillem Cucurull, David Esiobu, Jude Fernandes, Jeremy Fu, Wenyin Fu, Brian Fuller, Cynthia Gao, Vedanuj Goswami, Naman Goyal, Anthony Hartshorn, Saghar Hosseini, Rui Hou, Hakan Inan, Marcin Kardas, Viktor Kerkez, Madian Khabsa, Isabel Kloumann, Artem Korenev, Punit~Singh Koura, Marie-Anne Lachaux, Thibaut Lavril, Jenya Lee, Diana Liskovich, Yinghai Lu, Yuning Mao, Xavier Martinet, Todor Mihaylov, Pushkar Mishra, Igor Molybog, Yixin Nie, Andrew Poulton, Jeremy Reizenstein, Rashi Rungta, Kalyan Saladi, Alan Schelten, Ruan Silva, Eric~Michael Smith, Ranjan Subramanian, Xiaoqing~Ellen Tan, Binh Tang, Ross Taylor, Adina Williams, Jian~Xiang Kuan, Puxin Xu, Zheng Yan, Iliyan Zarov, Yuchen Zhang, Angela Fan, Melanie Kambadur, Sharan Narang, Aurelien Rodriguez, Robert Stojnic, Sergey Edunov, and Thomas
  Scialom.
\newblock Llama 2: Open foundation and fine-tuned chat models, 2023.

\bibitem[Xiao et~al.(2024)Xiao, Tang, Zuo, Guo, Yang, Tang, Fu, and Han]{duoattn}
Guangxuan Xiao, Jiaming Tang, Jingwei Zuo, Junxian Guo, Shang Yang, Haotian Tang, Yao Fu, and Song Han.
\newblock Duoattention: Efficient long-context {LLM} inference with retrieval and streaming heads.
\newblock \emph{CoRR}, abs/2410.10819, 2024.
\newblock \doi{10.48550/ARXIV.2410.10819}.
\newblock URL \url{https://doi.org/10.48550/arXiv.2410.10819}.

\bibitem[Yang et~al.(2022)Yang, Hu, Babuschkin, Sidor, Farhi, Pachocki, Liu, Chen, and Gao]{mup}
Greg Yang, Edward~J. Hu, Igor Babuschkin, Szymon Sidor, David Farhi, Jakub Pachocki, Xiaodong Liu, Weizhu Chen, and Jianfeng Gao.
\newblock Tensor programs v: Tuning large neural networks via zero-shot hyperparameter transfer.
\newblock In \emph{NeurIPS 2021}, March 2022.
\newblock URL \url{https://www.microsoft.com/en-us/research/publication/tuning-large-neural-networks-via-zero-shot-hyperparameter-transfer/}.

\bibitem[Yu et~al.(2022)Yu, Jeong, Kim, Kim, and Chun]{continuous}
Gyeong{-}In Yu, Joo~Seong Jeong, Geon{-}Woo Kim, Soojeong Kim, and Byung{-}Gon Chun.
\newblock Orca: {A} distributed serving system for transformer-based generative models.
\newblock In Marcos~K. Aguilera and Hakim Weatherspoon (eds.), \emph{16th {USENIX} Symposium on Operating Systems Design and Implementation, {OSDI} 2022, Carlsbad, CA, USA, July 11-13, 2022}, pp.\  521--538. {USENIX} Association, 2022.
\newblock URL \url{https://www.usenix.org/conference/osdi22/presentation/yu}.

\bibitem[Zaheer et~al.(2020)Zaheer, Guruganesh, Dubey, Ainslie, Alberti, Onta{\~{n}}{\'{o}}n, Pham, Ravula, Wang, Yang, and Ahmed]{bigbird}
Manzil Zaheer, Guru Guruganesh, Kumar~Avinava Dubey, Joshua Ainslie, Chris Alberti, Santiago Onta{\~{n}}{\'{o}}n, Philip Pham, Anirudh Ravula, Qifan Wang, Li~Yang, and Amr Ahmed.
\newblock Big bird: Transformers for longer sequences.
\newblock In Hugo Larochelle, Marc'Aurelio Ranzato, Raia Hadsell, Maria{-}Florina Balcan, and Hsuan{-}Tien Lin (eds.), \emph{Advances in Neural Information Processing Systems 33: Annual Conference on Neural Information Processing Systems 2020, NeurIPS 2020, December 6-12, 2020, virtual}, 2020.
\newblock URL \url{https://proceedings.neurips.cc/paper/2020/hash/c8512d142a2d849725f31a9a7a361ab9-Abstract.html}.

\bibitem[Zhang et~al.(2023)Zhang, Sheng, Zhou, Chen, Zheng, Cai, Song, Tian, R{\'{e}}, Barrett, Wang, and Chen]{h2o}
Zhenyu Zhang, Ying Sheng, Tianyi Zhou, Tianlong Chen, Lianmin Zheng, Ruisi Cai, Zhao Song, Yuandong Tian, Christopher R{\'{e}}, Clark~W. Barrett, Zhangyang Wang, and Beidi Chen.
\newblock H\({}_{\mbox{2}}\)o: Heavy-hitter oracle for efficient generative inference of large language models.
\newblock \emph{CoRR}, abs/2306.14048, 2023.
\newblock \doi{10.48550/arXiv.2306.14048}.
\newblock URL \url{https://doi.org/10.48550/arXiv.2306.14048}.

\bibitem[Zheng et~al.(2023)Zheng, Yin, Xie, Sun, Huang, Yu, Cao, Kozyrakis, Stoica, Gonzalez, et~al.]{sglang}
Lianmin Zheng, Liangsheng Yin, Zhiqiang Xie, Chuyue Sun, Jeff Huang, Cody~Hao Yu, Shiyi Cao, Christos Kozyrakis, Ion Stoica, Joseph~E Gonzalez, et~al.
\newblock Sglang: Efficient execution of structured language model programs, 2024.
\newblock \emph{URL https://arxiv. org/abs/2312.07104}, 2023.

\end{thebibliography}
\bibliographystyle{iclr2024_conference}

\appendix
\section{Appendix}
You may include other additional sections here.

\end{document}